%% file: PoFLSC.tex
\documentclass[conference]{IEEEtran}
\IEEEoverridecommandlockouts
\usepackage{cite}
\usepackage{amsmath,amssymb,amsfonts}
\usepackage{algorithmic}
\usepackage{graphicx}
\usepackage{textcomp}
\usepackage{xcolor}
\def\BibTeX{{\rm B\kern-.05em{\sc i\kern-.025em b}\kern-.08em
    T\kern-.1667em\lower.7ex\hbox{E}\kern-.125emX}}
\begin{document}

\title{Proof-of-Federated-Learning-Subchain: Free Partner Selection Subchain Based on Federated Learning}

\author{\IEEEauthorblockN{Boyang Li, Bingyu Shen, Qing Lu, Taeho Jung, Yiyu Shi}
\IEEEauthorblockA{\textit{Department of Computer Science and Engineering},
\textit{University of Notre Dame}, \\
Notre Dame, Indiana, U.S. \\
}
{\tt\small \{bli1,bshen,qlu2,tjung,yshi4\}@nd.edu}

}

\IEEEoverridecommandlockouts

\IEEEpubid{\makebox[\columnwidth]{978-8-3503-1019-1/23/\$31.00~\copyright2023 IEEE \hfill} \hspace{\columnsep}\makebox[\columnwidth]{ }}

\maketitle

\IEEEpubidadjcol

\begin{abstract}
The continuous thriving of the Blockchain society motivates research in novel designs of schemes supporting cryptocurrencies. Previously multiple Proof-of-Deep-Learning(PoDL) consensuses have been proposed to replace hashing with useful work such as deep learning model training tasks. The energy will be more efficiently used while maintaining the ledger. However deep learning models are problem-specific and can be extremely complex. Current PoDL consensuses still require much work to realize in the real world. In this paper, we proposed a novel consensus named Proof-of-Federated-Learning-Subchain(PoFLSC) to fill the gap. We applied a subchain to record the training, challenging, and auditing activities and emphasized the importance of valuable datasets in partner selection. We simulated 20 miners in the subchain to demonstrate the effectiveness of PoFLSC. When we reduce the pool size concerning the reservation priority order, the drop rate difference in the performance in different scenarios further exhibits that the miner with a higher Shapley Value (SV) will gain a better opportunity to be selected when the size of the subchain pool is limited. In the conducted experiments, the PoFLSC consensus supported the subchain manager to be aware of reservation priority and the core partition of contributors to establish and maintain a competitive subchain.
\end{abstract}


\input{body}

\bibliographystyle{IEEEtran} 
\bibliography{bib}

\end{document}

%% file: body.tex
\section{Introduction} \label{sec:1_intro}

The popularity of Blockchain in recent years has drawn an unprecedented amount of research attention in this field. Despite the advantages brought by decentralized mechanism, one of the main drawbacks of Blockchain, the tremendous energy cost, is causing increasing concern\cite{karmakar2021bitcoin}. Currently, most blockchains adopted hashing as workload which can only be solved through brute force. While hashing is secure and hard to crack, there's very limited additional value can be generated during this calculation process.

As a result, Proof-of-Deep-Learning (PoDL)\cite{chenli2019energy} was proposed in recent years to address the ``wasting energy'' issue, which replaced the hash algorithm with deep learning training tasks as the workload. Deep learning algorithms have been widely applied in various research areas such as computer vision and natural language processing. PoDL adopted the Proof-of-Work (PoW)\cite{chenli2019energy} and focused on designing pipelines and scheduling deep learning tasks to inherit the security properties. 
Successive works of PoDL chains such as DLchain\cite{chenli2020dlchain} and DLBC\cite{li2019dlbc} further improved security over the original design. 

However, training a deep learning (DL) model for a specific task is more complex compared with hashing.  In computer vision, the state-of-the-art neural network architectures in areas such as object detection\cite{yolov3} and image classification\cite{he2016deep} varied drastically depending on the details of the problem.  Moreover deep learning models are data driven\cite{solomatine2009data}. The quality of data used to train the model has a direct impact on the model's performance.

\begin{figure}
\centering
 \includegraphics[width=1\columnwidth]{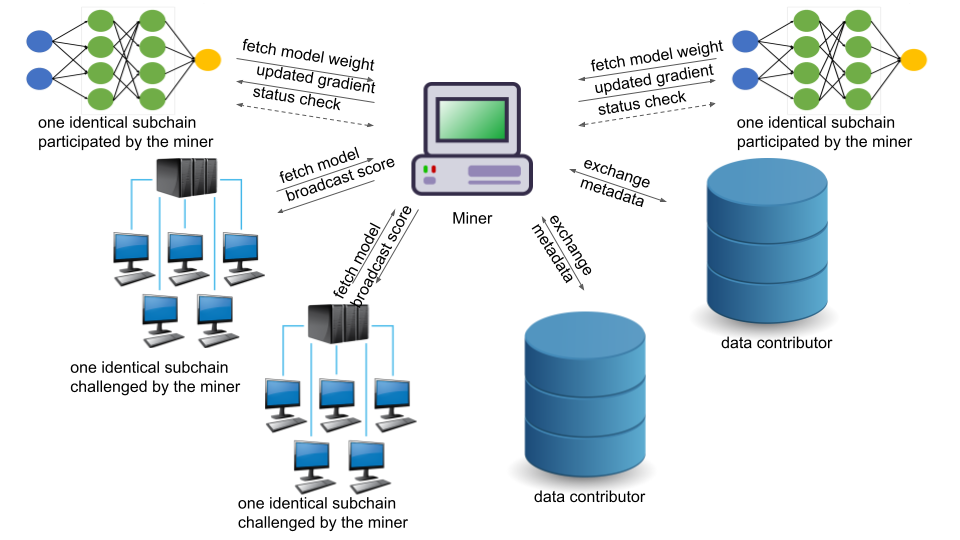}
 \caption{The interaction of a miner with others participants}
 \label{fig:miner}
\end{figure}

In this work, we introduce the novel consensus Proof-of-Federate-Learning-Subchain (PoFLSC) which is derived from PoDL. This is the first consensus that integrates the importance of the dataset into the task scheduling process among miners. In PoFLSC, miners are encouraged to collect and contribute their private datasets. Both the complexity of the model to train and the value of the dataset will be considered while miners choose mining partners and rank the priority of scheduled tasks. To enhance the security of PoFLSC, we adopted the challenge and witness verification mechanism from Helium\cite{garewal2020helium}. 
In the Fig. \ref{fig:miner}, it demonstrates the interaction between a miner and other contributor individuals/groups. Once the miner joins a subchain as a contributor, it maintains a ping-pong network communication to update the most result status with the subchain manager. Once a miner generates a challenge to a subchain, the miner fetches the model from the subchain and returns the performance of the model based on the private dataset of the miner. Amount visible data contributor, they share their metadata with each other to increase their visibility. The tasks of each roll will be introduced in the subsection \ref{sec:4_design:sub:tasks}

To conclude, a novel consensus named PoFLSC is proposed in this work. With the integration of data value and response time, miners will have the incentive to contribute their dataset and therefore made possible more complex deep learning tasks. Energy efficiency is improved compared with PoDL due to the diversity of the dataset introduced by miners. Model performance will also benefit from the increased scale of the dataset.



\section{Background} \label{sec:2_back}

\subsection{Related work}
Previously, the PoDL\cite{chenli2019energy} consensus mainly utilized the computation capability of miners, and the tasks publisher release the training tasks and the training data. In the PoNAS\cite{li2021mining}, the tasks publisher provides training data and relatively flexible training tasks. The target task is to search a neural architecture network, thus the actual training tasks can be different from each other.   

Federated learning(FL) is proposed as an efficient deep learning method suitable for decentralized data\cite{mcmahan2017communication}. Using FL, millions of users can train a model together with local data in their devices, and only gradients will be uploaded and aggregated to update the shared model's weights. Some advantages are essential using FL compared with traditional deep learning training. Firstly, the data remains private during the entire training process. Secondly, the hardware requirement of the user device is minimum and network resources are saved.

Furthermore, the combination of blockchain and FL resolved some of the existing drawbacks in FL such as centralized server, robust network communication, and lack of incentive\cite{nguyen2021federated}. FL-blockchains\cite{8733825} removed the central server role and minimized the impact of remote devices failures. More than that, FL-blockchains naturally motivated devices to participate and contribute to the chain.



In our PoFLSC, because all miners will have a strong incentive to contribute their dataset, the training process will less likely suffer the problem of data scarcity. As a guarantee, it's important to fairly reward each miner considering the contribution of each dataset. Wang et al\cite{9006179} proposed to use Shapley Values (SV) to measure the contributions of participants in FL. Influence estimation for each party in horizontal FL and Shapley estimation for individual feature value are considered in their work. To alleviate the calculation in this work, Ghorbani et al\cite{ghorbani2019data} proposed another Shapley-Value-based evaluation method to quantify the value of each training datum to the trained model's performance which is named Truncated Monte Carlo Shapley.



\subsection{Participants}
1) Miners are the machines that join the decentralized network and contribute the resource for crypt-currency reward. 
In PoFLSC, the ability of DL model training, hosting the pool manager, proxy, high-quality data collecting are all crucial to earn rewards. 

2) Full nodes will perform three types of checks. 
i) All nodes can behave as Type One full nodes and it will record and check all blocks and transactions;
ii) As Type Two full node, each data contributor will generate challenges periodically to test the DL model performance of all its visible subchain;
iii) The Type Three full node will audit the training procedure by repeating it. 

3) Task publisher will provide a certain DL model architecture and the sample dataset. In PoFLSC, the public accessible dataset is less scarce and miners will merge the public dataset into the training if the SV of the dataset is higher than the selection threshold. The majority of training data are private datasets from data contributors.


\subsection{Assumptions}

This work is based on three assumptions. 

1) The dataset is clean. 
In this work, we only consider the quality of datasets. We assume no adversarial attack, poison attack, or miss label issue happens in any dataset. All mentioned attacking cases has been evaluated in related work \cite{ghorbani2019data}. These attack mechanism will reduce the SV of the miner and potentially reduce the performance of the subchain. But the subchain manager will less likely to reserve a miner with low SV dataset. Therefore, the miner with mentioned attacking cases will not be selected. 


2) The size of a dataset from each data contributor is the same and the size of each sample is the same. With this assumption, we can avoid discussing whether a dataset with large in size and medium in quality is more valuable than a dataset with small in size and high in quality. 

3) Each block will finish one task and the tasks between two blocks are independent. Therefore, we will not discuss the case if data contributors wish to hide part of their private dataset in training. Within one block, all the sharing activity will be recorded and data contributors share the high-quality dataset to their core pool partners.



\section{Design} \label{sec:4_design}

\subsection{Overview}

The design of PoFLSC inherits the PoDL consensus and proposes modifications to improve the effectiveness when facing complex scenarios. 
PoFLSC allows miners to select/reject partners, challenge/verify the performance and configuration of each other. Specifically, PoFLSC is a free market for miners to select partners by measuring the response time and data value of other miners. This dynamic free market provides incentives for miners to contribute valuable data or a dataset from the minority group. 


To train a model is a more complex task than hash algorithm workload. It requests high performance of network, computation, storage, data, and model design. PoNAS\cite{li2021mining} adopted PoDL design and proposed a mining pool solution on top of NAS workload. The pool manager scheduled strong miners to find potential neural network structures and weak miners to fine-tune the given model. The manager will assign a random sub search space within the full space for each strong miner. This training process among different miners is independent, therefore the performance of a single searching task for a miner depends on the neural network architecture, searching space, training data, etc. The performance of partners in the pool cannot hold back the performance of any other member in the pool. When we adopt this mining pool strategy to distribute deep learning framework, an intuitive method is to split the model or training data of deep learning training tasks and assign it as a sub-task to miners. With this strategy, the miners may cumber the overall performance of the pool if the miners are slow in computation power or network speed.  

In PoFLSC, response time and value of data are key factors for miners to select partners. The general idea is to encourage miners work with as many partners as possible within one sub-block time of the subchain, then rank the priority of partners according to their data SV. It is allowed that every miner can contribute to many subchain. In general, one main-block finishes one DL training task and one FL global communication round is finished within one sub-block. All miners experience fours phase including the initial phase, core pool establish phase, secondary pool establish phase, and verification phase. 

\subsection{Miner tasks} \label{sec:4_design:sub:tasks}

1) Training: 
With given neural network architecture and local dataset or partner shared dataset, the training workforce train DL models for certain local epoch and submit the updated gradients to the host or other miners. 

2) Hosting the pool manager: 
Besides strong computation power, the host is the server to manage multiple other miners. 
Within the core pool, the host is selected with the conditions: 
i) long term reliability exceeds core pool average value; 
ii) among all qualified candidacies, the selected host must have the shortest response time. The host also aggregates the gradients of all updated gradients from others member of the core pool and distributes the average global gradients to each member. 

3) Proxy: 
The proxy can forward data or requests. With strong proxy involve, the core pool network performance of bandwidth saving and speeds will be improved. 

4) Data Contributor:
Data contributors will collect private datasets for DL training tasks and the owner of the high-quality dataset will be the preferred partner in the PoFLSC. 
Because each miner is allowed to contribute in multiple subchains if the miner is able to submit the updated gradients within one sub-block time of the subchain, the high-quality data contributor will earn more opportunity to be the final winner. 

\subsection{Subchain Structure}

In PoFLSC, each block will finish one target task and each block will be divided into multiple sub-blocks. The sub-block time is predefined when releasing the training tasks and it is longer than multiple local epoch time for slow miners. A relatively short sub-block time will limit the number of miners to contribute to the final DL model, while increasing the global communication rounds. With relatively longer sub-block time, it encourages more miners to join the competition and it encourages miners to participate in multiple subchains. Working with multiple subchains in parallel will increase the opportunity to win. Therefore, the miner or a core pool with a valuable dataset or shorter response time will wish to work with multiple subchains and they will increase their winning chance by upgrading to better hardware or collecting valuable data.  

The main structure of subchain is similar to PoDL \cite{li2019dlbc} which includes a block head and a block body. The block head structure remains the same as in PoDL. The first sub-block head of the current block is the hash of the previous block head. Among all visible subchains, only one winner subchain will write their sub-block information as the block information for the current block. After it runs into the fourth phase, a subchain becomes a candidate once the number of audits and challenges exceed the threshold, respectively. The winner subchain is the one with the best accuracy among all candidates. 
All qualified subchains should receive enough audits and challenges as confirmation. 

\begin{figure}
\centering
 \includegraphics[width=1\columnwidth]{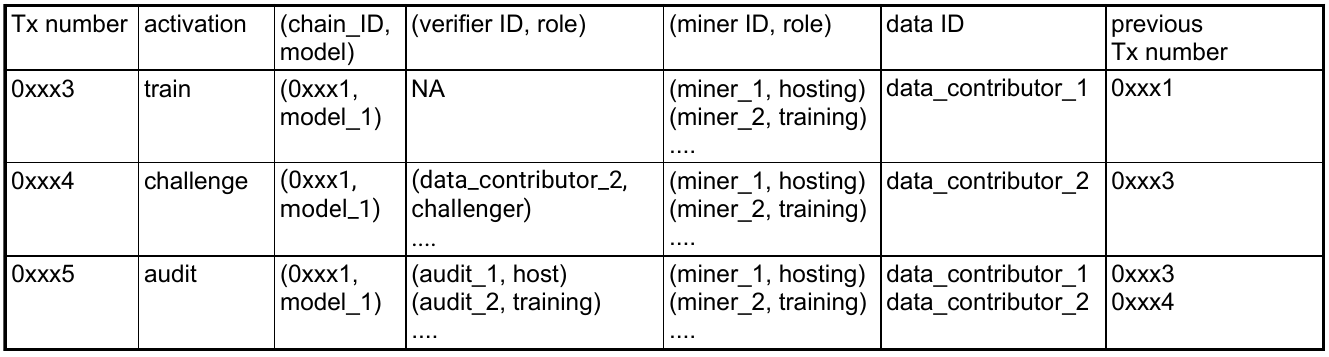}
 \caption{A sample of activation transaction}
 \label{fig:tx}
\end{figure}

The block body records transaction ledger and activation transaction as shown in Fig. \ref{fig:tx}
The transaction ledger remains the same as in PoDL. The activation transaction includes information on training, challenges, and audits. All the challenge results will be the model performance. Auditing will further verify the training procedure. In the activation transaction, it records transaction number, activation type, chain ID and model pair, verifier ID and role pair, miner ID and role pair, data ID, and previous dependency transaction number. 

\subsection{Four phase of intervals}

There are four phase intervals within one block. Each interval lasts one or more sub-block time. 
There is no certain time limit for all subchains to follow. But within one certain subchain, all members follow the same guide. The concept of this phase is for all participants to be aware of the status of the training procedure, thus it will receive resources for the different types of tasks. 

In general, once the miner selects partners to establish a core pool or secondary pool. The miners' pool will adopt federated learning framework to train DL model together. Each node will run the required number of local epoch and submit the gradients to the host manager. The host manager will update the server average gradients and distribute the updated gradients to each following miner. One sub-block time is one global communication round.  

1) The first phase is the initial phase to select partners to form core pools. 
Each miner maintains an event queue for an upcoming response time check and the time for the request in the queue is not part of response time. The response time includes the time for training the task and transferring data. Each miner also checks all visible miners for response time and maintains a list of partner candidates. The list adds a new partner if the sum of the response time of all candidates is shorter than one sub-block time or the response time of the new partner is shorter than the longest partner in the list. When the total response time of all partners is longer than one sub-block time, it removes the partner with the longest response time in the list. 
Between each pair of partners, they share their partner list starting from the partner with the lowest response time.
Once a partner is a common partner in the pair, it establishes the core pool. All members of the core pool propose one candidate if the response time is shortest among all unconfirmed candidates in their list. Once any proposed candidate is not on the local list, the miner will raise reject and the candidate selection will stop. If the number of all confirmed candidates is bigger than the threshold number, the core pool 
will be established, otherwise, the core pool will be demolished. All participants of each core pool will maintain one subchain. 


2) Once the core pool is established, the phase moves on to the second phase. All miners start training tasks and evaluate SV of the partners in each participated subchains. Based on the SV, the miner will rank the priority of these subchains in the local task schedule. The time for the initial phase is much short than one sub-block time and we consider all subchain will keep the same number of sub-block by the end because all subchains start to generate subblocks within the first sub-block and the length of each sub-block time is the same. 
Here, each subchain adopts a synchronous federated learning framework in the second phase. The pool manager nodes collect all unconfirmed candidates from each member and repeat the same algorithm in phase one to select partnerships with other pool managers. But all selection behavior happens among the pool manager of each unconfirmed candidate. 

3) Once the partnership among multiple pool managers is confirmed, it establishes the secondary pool, and the phase moves onto the third. Each subchain will split and merge. For each subchain, the number of branches it split equals the number of partnerships. One branch of all subchain within one partnership merge into one subchain. In the third phase, the workforce nodes receive a global average gradients from their manager and continue training on the updated gradients. The manager adopts an asynchronous federated learning framework which allows the manager to boost the response time with faster miners, thus it helps the core pool increase the priority ranking in other schedules. The manager nodes update response time and evaluate SV with other manager nodes in partnership. 


4) A subchain runs into the fourth phase when a secondary pool achieves both requirements: i) received sufficient audits and challenge, ii) qualified performance of workload. This subchain becomes the candidate of the final winner and it receives more challenges and audits from others.


\subsection{Verification}

Full nodes will perform three types of checks. 

i) All nodes can behave as Type One full nodes and it will record and check all blocks and transactions. In addition, all challenges and verification activation will be recorded as transactions too. 
The Type One results will be transaction confirmation and transaction reliability records of a single node. 

ii) As Type Two full node, each data contributor will generate challenges periodically to test the DL model performance of all its visible subchain.
The Type Two full node is limited to only generating one challenge set in one period of time. The challenge set is multiple random subsets of its private dataset. One subchain will receive only one challenge from one Type Two challenge in one period.
The Type Two results will be the DL model performance and pool performance record. 

iii) The Type Three full node will audit the training procedure by repeating it. 
Auditing of the training procedure is a heavy workload and it will be finished with multiple nodes in a group.
The results of auditing will be the comprehensive record of a pool and all participants in their performance and reliability. 

\begin{figure}
\centering
 \includegraphics[width=1\columnwidth]{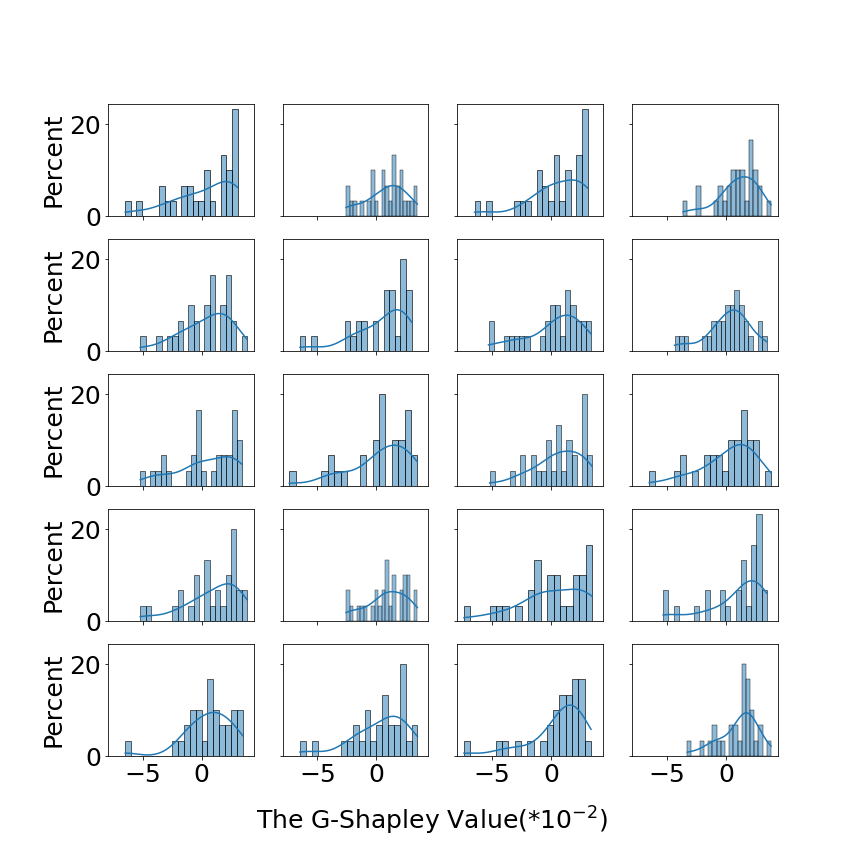}
         \caption{The histogram of 20 miners with the G-Shapley Value method.} 
         \label{fig:exp:histogram_G}
\end{figure}

\begin{figure}
\centering
         \includegraphics[width=1\columnwidth]{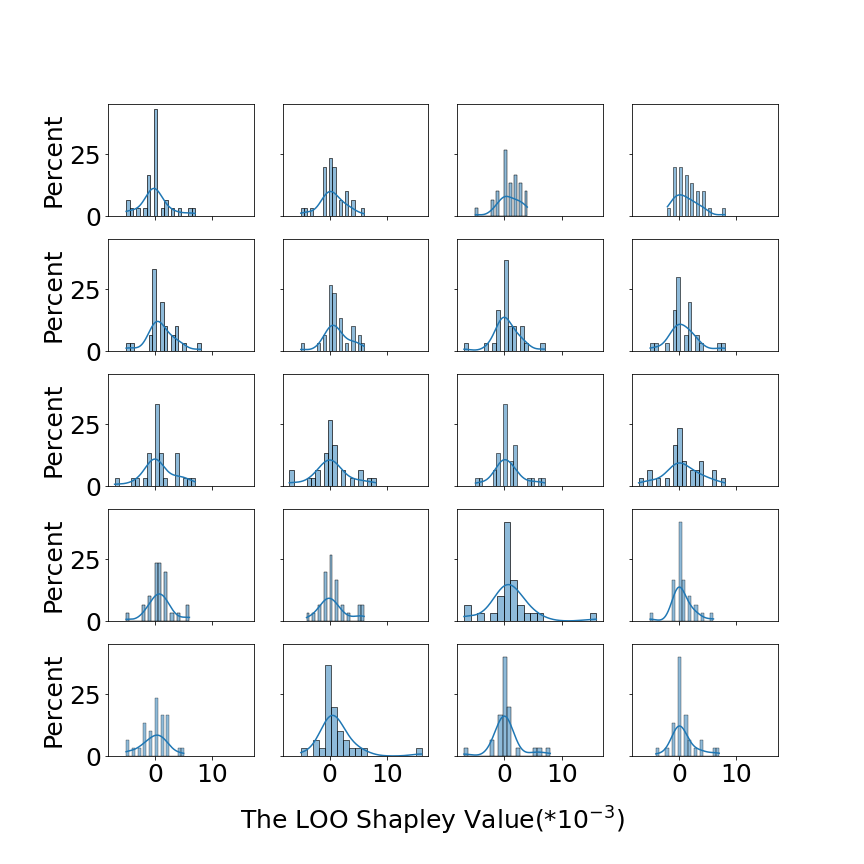} 
         \caption{The histogram of 20 miners with the LOO Shapley Value method.} 
         \label{fig:exp:histogram_loo}
\end{figure}

\section{Experiment} \label{sec:5_exper}

\subsection{Experimental Setup}
In the experiment, we evaluated the effectiveness of PoFLSC. When the training process is finished in the main chain, all participants agree that the subchain with the best performance model will write the block and initial the next block. 
The PoFLSC consensus is proposed to augment the functionality within each block.   
In the first phase, it will initiate the core pool based on the response time. 
In the secondary phase, the core pool started training and evaluate the SV of each miner. Because the pool need to finish at least one global epoch, the manager will measure the reservation priorities of each miner. When the resource is limit that the pool manager cannot afford handling all miners, the manager will firstly reserve the miner with the highest ranking in reservation priority queue, therefore the pool manager guarantees that the core pool will finish global epoch within one subblock time. It is important that every subchain can meet this commitment, thus the subchain can continue to join the secondary pool and final competition.

We simulated 100 miners to verify the effectiveness. 
The training task is MNIST \cite{lecun2010mnist} digits handwriting recognition and the DL model is of three 2D convolution layers and two fully connected layers. 
Each miner randomly select 30 samples from all samples. 
The response time between each pair of miners is given and fixed.
The statistics of the response time is simulated to follow the Gaussian distribution. 
The demonstration subchain selected top 20 miners as the members of the core pool.
In the experiment, the SV of all miners are calculated in two different methods. 
In the Fig. \ref{fig:exp:histogram_G} and \ref{fig:exp:histogram_loo}, it shows the histogram of the SV of 20 miners. The Fig. \ref{fig:exp:histogram_G} is for G-Shapley Value method and The Fig. \ref{fig:exp:histogram_loo} is for LOO Shapley Value method

\begin{figure}
\centering
         \includegraphics[width=1\columnwidth]{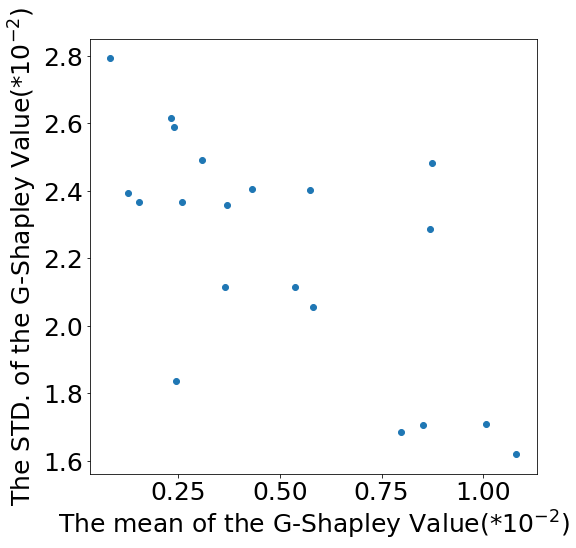}
         \caption{The statistics results of the G-Shapley Value of the subchain. Each point represents a member.}
         \label{fig:exp:stat_G}
\end{figure}

\begin{figure}
\centering
         \includegraphics[width=1\columnwidth]{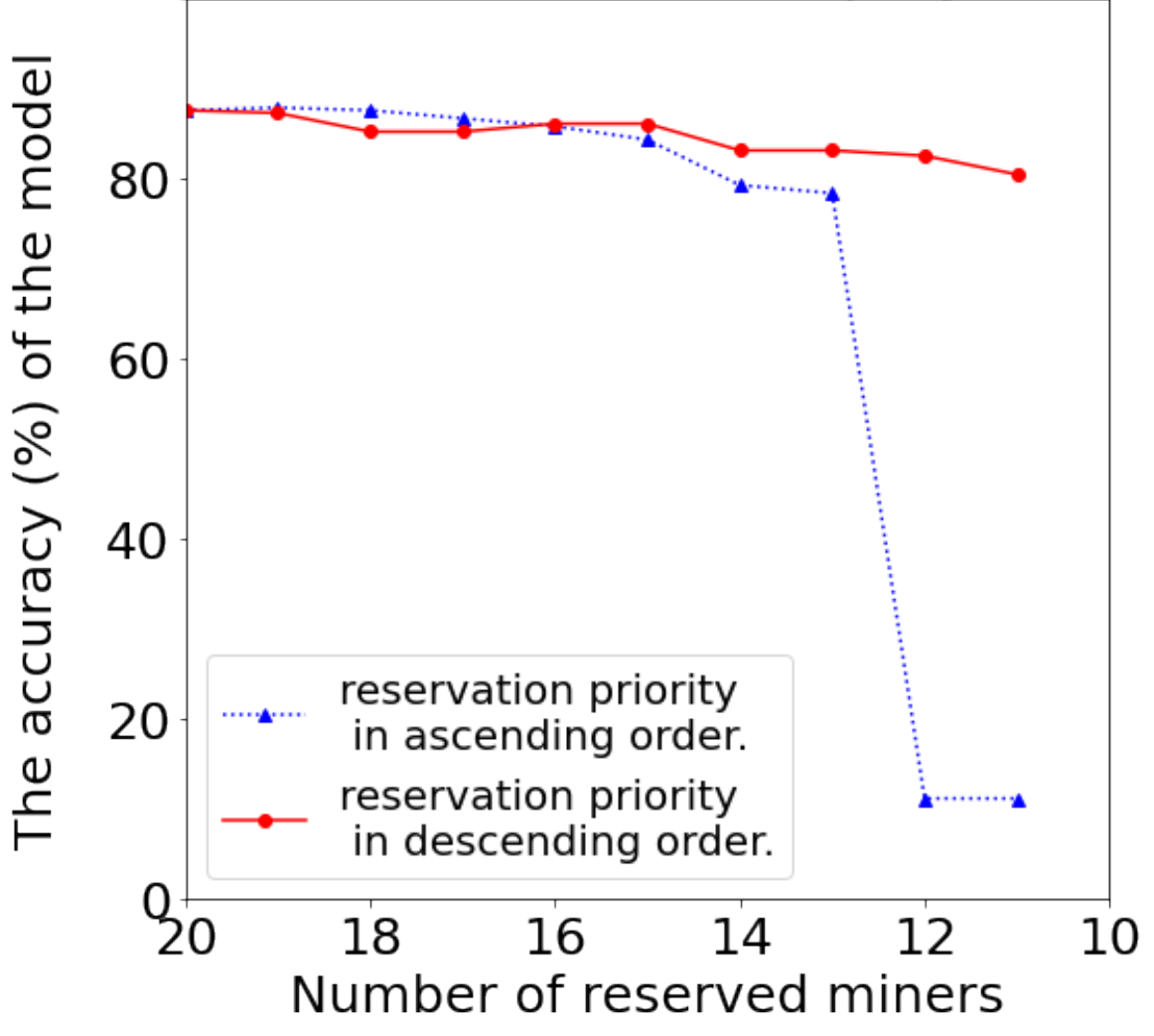}
         \caption{The performance of models in two different reservation priority order shows the effectiveness.}
         \label{fig:exp:perf_G}
\end{figure}


\begin{figure}
\centering
         \includegraphics[width=1\columnwidth]{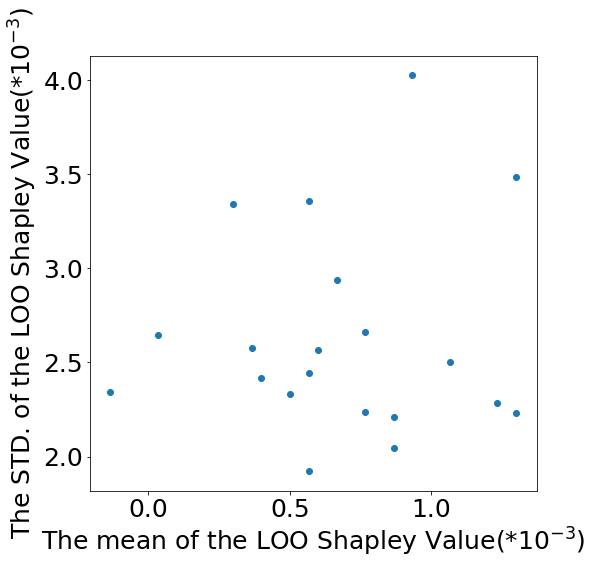}
         \caption{The statistics results of the LOO Shapley Value of the subchain. Each point represents a member.}
         \label{fig:exp:stat_LOO}
\end{figure}

\begin{figure}
\centering
         \includegraphics[width=1\columnwidth]{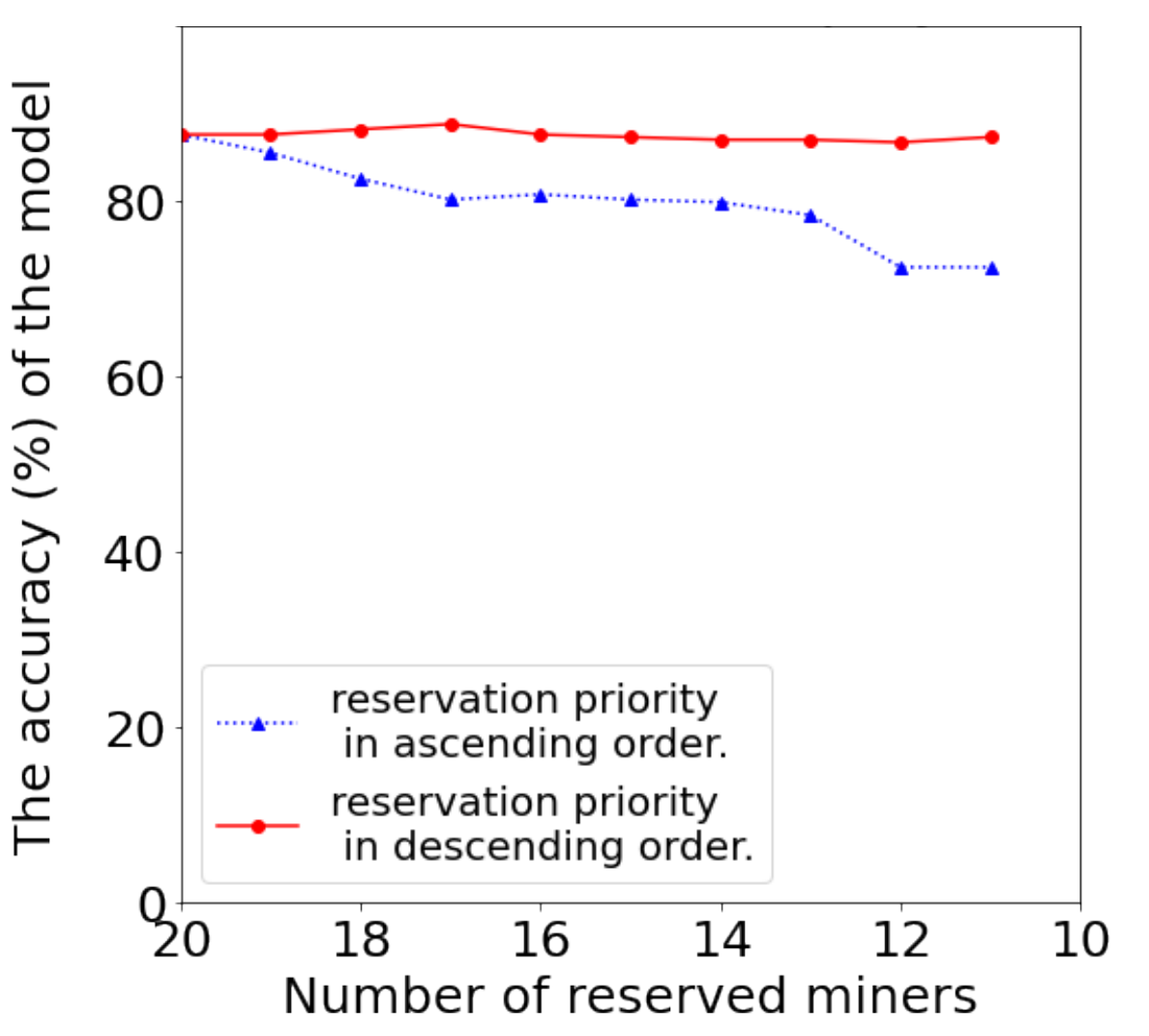}
         \caption{The performance of models in two different reservation priority order shows the effectiveness of PoFLSC.}
         \label{fig:exp:perf_LOO}
\end{figure}

     


\subsection{Performance Evaluation}


In the Fig. \ref{fig:exp:stat_G} and \ref{fig:exp:stat_LOO}, it shows mean and STD. of member where the x coordinate represents the mean of the SV of each member and the y coordinate represents the STD. of the SV of each member. 
Here, we compared the performance differences when we reserved members in ascending order and descending order in the Fig. \ref{fig:exp:perf_G} and \ref{fig:exp:perf_LOO}. The solid line represents the descending order which means we will reserve the candidate with the highest SV first when we start adding more member, and it also means we will kick the member with the lowest SV first when we start removing member. The dotted line shows the results of comparative experiment where the manager will reserve the candidate with the lowest SV first and remove the member with the highest SV first.



Due to the limitation of computation resources, we limited the size of the subchain pool to 20 in our simulation. 
When we gradually reduce the pool size, we observe the performance drop in all cases. The drop rate of the solid line is more gentle than the dotted line. When the pool size drops to 60\%, the performance drops below 20\% in Fig. \ref{fig:exp:perf_G}. So the top 40\% of members are the key contributors to be reserved to maintain the subchain in this experiment. In practice, the value of this threshold can be various in different tasks and datasets, but it is important for the manager to be aware of reservation priority and the core partition of contributors to maintain a healthy subchain. 
In Fig. \ref{fig:exp:perf_G} and \ref{fig:exp:perf_LOO}, the drop rate of the performance means the relevance between the SV and the value of the data for this task. 
The drop rate difference between the dotted lines in Fig. \ref{fig:exp:perf_G} and \ref{fig:exp:perf_LOO} shows that the relevance is higher with G-shapley method than it is with LOO method. 
This experiment shows the effectiveness of this novel PoFLSC consensus and it demonstrates the miner with higher SV will achieve a better opportunity to be selected when the size of the subchain pool is limited.
The reservation priority order based on SV is helpful for a subchain manager to select candidates and establish a competitive subchain.   




\section{Discussion} \label{sec:6_disc}

Previously, the PoDL consensus mainly utilized the computation capability of miners for deep learning training. The training tasks and the training data are given by the tasks publisher.
In the PoNAS, the tasks publisher provided training data and relatively flexible training tasks.
The target task is to search a neural architecture network, thus the actual training tasks can be different from each other.
In this work, we considered the complexity of training a deep learning model and proposed the PoFLSC. Here, it emphasizes the importance of training data in DL models. In addition, the design also concerns the response time of miners, verification between miners. The response time can be expended to computation power of miner and specification of network infrastructure, etc. The verification mechanism further strengthens the security of the system.


Because DL models are data-driven, it is necessary to evaluate the value of the dataset in this novel consensus.
To my best knowledge, we are the first to adopt SV into novel consensus and further evaluate the effectiveness of PoFLSC.
The value of datasets in the consensus also improves blockchain security. For example, if there is no mechanism to evaluate the value of each dataset, the honest miners will have less motivation to collect private datasets or generate synthetic datasets, while the attackers have strong motivation to collect or generate high-quality private datasets to improve model performance and win the competition. 
Once the consensus requests private datasets and evaluates the value of each private dataset, all miners will have the motivation to collect high-quality data. As a result, more effective and economical methods of data collection are expected to be proposed. For example, existing works \cite{shen2021automatic,bhandari2018procedural} have proved to be capable of generating synthetic training data for computer vision tasks utilizing 3D simulation techniques. Besides helping to create incentives for data collection, SV also contributes to detecting low-quality datasets, miss label samples, and adversarial attacks \cite{wang2021unified}.  


